%% file: main.tex
\documentclass[11pt,a4paper]{article}
\usepackage[hyperref]{acl2018}
\usepackage{times}
\usepackage{latexsym}
\usepackage{blindtext}
\usepackage{amsmath}
\usepackage{tikz}
\usetikzlibrary{%
  trees,%
  shapes,%
  arrows,%
  positioning,%
  calc,%
  automata%
}
\usepackage{tikz-dependency}
\usepackage{pgfplots}
\pgfplotsset{compat=1.14}
\usepackage{multirow}
\usepackage{hhline}
\usepackage{booktabs}
\usepackage{enumitem}
\usepackage{graphicx}

\usepackage{url}

\aclfinalcopy 
\def\aclpaperid{270} 

\newcommand{\smalltt}[1]{{\small\texttt{#1}}}
\newcommand{\smallsf}[1]{{\small\textsf{#1}}}
\newcommand{\smallsc}[1]{{\small\textsc{#1}}}

\newcommand{\specialcell}[2][c]{%
  \begin{tabular}[#1]{@{}c@{}}#2\end{tabular}}
  
\definecolor{my_green}{rgb}{0, 0.6, 0.5}
\definecolor{my_orange}{rgb}{0.9, 0.6, 0}
\definecolor{my_purple}{rgb}{0.8, 0.6, 0.7}
\definecolor{my_blue}{rgb}{0, 0.45, 0.7}

\title{Paraphrase to Explicate:\\
Revealing Implicit Noun-Compound Relations}

\author{Vered Shwartz ~~~~~~~~~~~~~~~~~~~~~~~~~~~~~~~~~~~~~~~ Ido Dagan\\
   Computer Science Department, Bar-Ilan University, Ramat-Gan, Israel\\
   {\tt vered1986@gmail.com	~~~~~~	\tt dagan@cs.biu.ac.il }}

\date{}

\begin{document}
\maketitle
\begin{abstract}
  Revealing the implicit semantic relation between the constituents of a noun-compound is important for many NLP applications. It has been addressed in the literature either as a classification task to a set of pre-defined relations or by producing free text paraphrases explicating the relations. Most existing paraphrasing methods lack the ability to generalize, and have a hard time interpreting infrequent or new noun-compounds.
We propose a neural model that generalizes better by representing paraphrases in a continuous space,  generalizing for both unseen noun-compounds and rare paraphrases. Our model helps improving performance on both the noun-compound paraphrasing and classification tasks.
\end{abstract}

\section{Introduction}
\label{sec:intro}
\input{introduction}

\section{Background}
\label{sec:background}
\input{background}

\section{Paraphrasing Model}
\label{sec:model}
\input{model}

\section{Qualitative Analysis}
\label{sec:qualitative_analysis}
\input{qualitative_analysis}

\section{Evaluation: Noun-Compound Interpretation Tasks}
\label{sec:tasks}
For quantitative evaluation we employ our model for two noun-compound interpretation tasks. The main evaluation is on retrieving and ranking paraphrases (\S\ref{sec:main_paraphrasing}). For the sake of completeness, we also evaluate the model on classification to a fixed inventory of relations (\S\ref{sec:main_classification}), although it wasn't designed for this task.

\input{paraphrasing}
\input{classification}

\section{Compositionality Analysis}
\label{sec:compositionality}
\input{compositionality}

\section{Conclusion}
\label{sec:conclusion}

We presented a new semi-supervised model for noun-compound paraphrasing. The model differs from previous models by being trained to predict both a paraphrase given a noun-compound, and a missing constituent given the paraphrase and the other constituent. This results in better generalization abilities, leading to improved performance in two noun-compound interpretation tasks. In the future, we plan to take generalization one step further, and explore the possibility to use the biLSTM for generating completely new paraphrase templates unseen during training. 

\section*{Acknowledgments}

This work was supported in part by an Intel ICRI-CI grant, the Israel Science Foundation grant 1951/17, the German Research Foundation through the German-Israeli Project Cooperation (DIP, grant DA 1600/1-1), and Theo Hoffenberg. Vered is also supported by the Clore Scholars Programme (2017), and the AI2 Key Scientific Challenges Program (2017).

\bibliography{bib}
\bibliographystyle{acl_natbib}

\end{document}

%% file: introduction.tex
Noun-compounds hold an implicit semantic relation between their constituents. For example, a `birthday cake' is a cake \textit{eaten on} a birthday, while `apple cake' is a cake \textit{made of} apples. Interpreting noun-compounds by explicating the relationship is beneficial for many natural language understanding tasks, especially given the prevalence of noun-compounds in English \cite{nakov2013interpretation}. 

The interpretation of noun-compounds has been addressed in the literature either by classifying them to a fixed inventory of ontological relationships \cite[e.g.][]{nastase2003exploring} or by generating various free text paraphrases that describe the relation in a more expressive manner \cite[e.g.][]{S13-2025}. 

\pagebreak

Methods dedicated to paraphrasing noun-compounds usually rely on corpus co-occurrences of the compound's constituents as a source of explicit relation paraphrases \cite[e.g.][]{wubben2010uvt,versley2013sfs}. Such methods are unable to generalize for unseen noun-compounds. Yet, most noun-compounds are very infrequent in text \cite{kim2007interpreting}, and humans easily interpret the meaning of a new noun-compound by generalizing existing knowledge. For example, consider interpreting \textit{parsley cake} as a cake made of parsley vs. \textit{resignation cake} as a cake eaten to celebrate quitting an unpleasant job. 

We follow the paraphrasing approach and propose a semi-supervised model for paraphrasing noun-compounds. Differently from previous methods, we train the model to predict either a paraphrase   expressing the semantic relation of a noun-compound (predicting `[w$_2$] made of [w$_1$]' given `apple cake'), or a missing constituent given a combination of paraphrase and noun-compound (predicting `apple' given `cake made of [w$_1$]'). Constituents and paraphrase templates are represented as continuous vectors, and semantically-similar paraphrase templates are embedded in proximity, enabling better generalization. Interpreting `parsley cake' effectively reduces to identifying paraphrase templates whose ``selectional preferences'' \cite{pantel-EtAl:2007:main} on each constituent fit `parsley' and `cake'.

A qualitative analysis of the model shows that the top ranked paraphrases retrieved for each noun-compound are plausible even when the constituents never co-occur (Section~\ref{sec:qualitative_analysis}). We evaluate our model on both the paraphrasing and the classification tasks (Section~\ref{sec:tasks}). On both tasks, the model's ability to generalize leads to improved performance in challenging evaluation settings.\footnote{The code is available at \scriptsize\url{github.com/vered1986/panic}}

\pagebreak

%% file: background.tex
\subsection{Noun-compound Classification}
\label{sec:bg_nc_classification}

Noun-compound classification is the task concerned with automatically determining the semantic relation that holds between the constituents of a noun-compound, taken from a set of pre-defined relations. 

Early work on the task leveraged information derived from lexical resources and corpora \cite[e.g.][]{girju:2007:ACLMain,oseaghdha-copestake:2009:EACL,tratz-hovy:2010:ACL}. More recent work broke the task into two steps: in the first step, a noun-compound representation is learned from the distributional representation of the constituent words \cite[e.g.][]{mitchell2010composition,zanzotto2010estimating,D12-1110}. In the second step, the noun-compound representations are used as feature vectors for classification \cite[e.g.][]{dima2015automatic,W16-1604}. 

The datasets for this task differ in size, number of relations and granularity level \cite[e.g.][]{nastase2003exploring,kim2007interpreting,tratz-hovy:2010:ACL}. The decision on the relation inventory is somewhat arbitrary, and subsequently, the inter-annotator agreement is relatively low \cite{kim2007interpreting}. Specifically, a noun-compound may fit into more than one relation: for instance, in \newcite{tratz2011semantically}, \textit{business zone} is labeled as \textsc{contained} (zone contains business), although it could also be labeled as \textsc{purpose} (zone whose purpose is business).

\subsection{Noun-compound Paraphrasing}
\label{sec:bg_nc_paraphrasing}

As an alternative to the strict classification to pre-defined relation classes, \newcite{nakov2006using} suggested that the semantics of a noun-compound could be expressed with multiple prepositional and verbal paraphrases. For example, \textit{apple cake} is a cake \emph{from}, \emph{made of}, or which \emph{contains} apples. 

The suggestion was embraced and resulted in two SemEval tasks. SemEval 2010 task 9 \cite{butnariu-EtAl:2009:SEW} provided a list of plausible human-written paraphrases for each noun-compound, and systems had to rank them with the goal of high correlation with human judgments. In SemEval 2013 task 4 \cite{S13-2025}, systems were expected to provide a ranked list of paraphrases extracted from free text.

Various approaches were proposed for this task. Most approaches start with a pre-processing step of extracting joint occurrences of the constituents from a corpus to generate a list of candidate paraphrases. Unsupervised methods apply information extraction techniques to find and rank the most meaningful paraphrases 
\cite{D11-1060,XAVIER14.125,N15-1037,pavlick-pasca:2017:Long}, while supervised approaches learn to rank paraphrases using various features such as co-occurrence counts \cite{wubben2010uvt,li2010ucd,surtani2013iiit,versley2013sfs} or the distributional representations of the noun-compounds \cite{vandecruys-afantenos-muller:2013:SemEval-20132}. 

One of the challenges of this approach is the ability to generalize. If one assumes that sufficient paraphrases for all noun-compounds appear in the corpus, the problem reduces to ranking the existing paraphrases. It is more likely, however, that some noun-compounds do not have any paraphrases in the corpus or have just a few. The approach of \newcite{vandecruys-afantenos-muller:2013:SemEval-20132} somewhat generalizes for unseen noun-compounds. They represented each noun-compound using a compositional distributional vector \cite{mitchell2010composition} and used it to predict paraphrases from the corpus. Similar noun-compounds are expected to have similar distributional representations and therefore yield the same paraphrases. For example, if the corpus does not contain paraphrases for \textit{plastic spoon}, the model may predict the paraphrases of a similar compound such as \textit{steel knife}. 

In terms of sharing information between semantically-similar paraphrases, \newcite{nulty2010ucd} and \newcite{surtani2013iiit} learned ``is-a'' relations between paraphrases from the co-occurrences of various paraphrases with each other. For example, the specific `[w$_2$] extracted from [w$_1$]' template (e.g. in the context of \textit{olive oil}) generalizes to `[w$_2$] made from [w$_1$]'. One of the drawbacks of these systems is that they favor more frequent paraphrases, which may co-occur with a wide variety of more specific paraphrases.

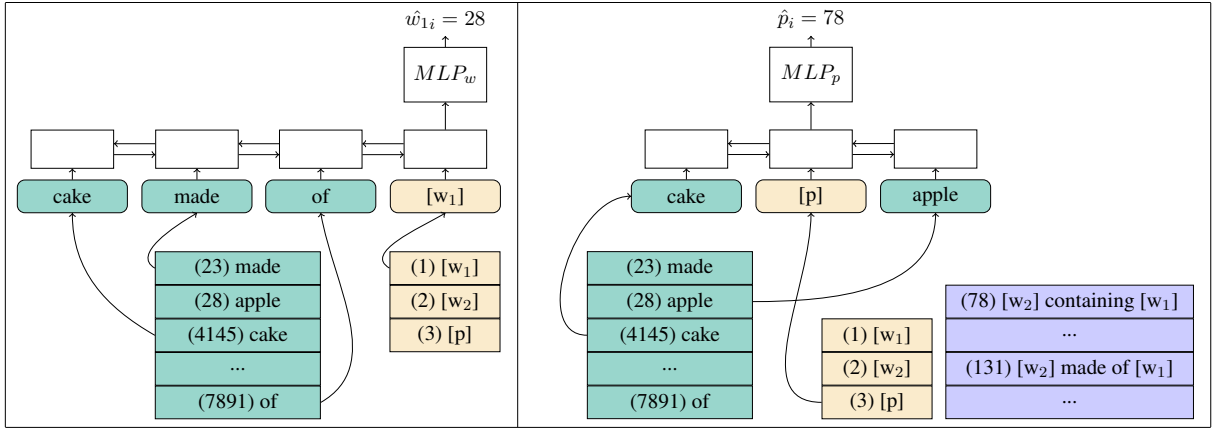
\begin{figure*}[!t]
\centering
\input{figures/model}
\vspace*{-14pt}
\caption{An illustration of the model predictions for $w_1$ and $p$ given the triplet \textit{(cake, made of, apple)}. The model predicts each component given the encoding of the other two components, successfully predicting `\textit{apple}' given `\textit{cake made of} [w$_1$]', while predicting `[w$_2$] \textit{containing} [w$_1$]' for `\textit{cake} [p] \textit{apple}'.} 
\vspace*{-12pt}
\label{fig:model}
\end{figure*}

\subsection{Noun-compounds in other Tasks}
\label{sec:bg_paraphrasing_oie}

Noun-compound paraphrasing may be considered as a subtask of the general paraphrasing task, whose goal is to generate, given a text fragment, additional texts with the same meaning. However, 
general paraphrasing methods do not guarantee to explicate implicit information conveyed in the original text. Moreover, the most notable source for extracting paraphrases is multiple translations of the same text \cite{P01-1008,N13-1092,mallinson-sennrich-lapata:2017:EACLlong}. If a certain concept can be described by an English noun-compound, it is unlikely that a translator chose to translate its  foreign language equivalent to an explicit paraphrase instead.

Another related task is Open Information Extraction \cite{etzioni2008open}, whose goal is to extract relational tuples from text. Most system focus on extracting verb-mediated relations, and the few exceptions that addressed noun-compounds provided partial solutions. \newcite{pal-:2016:W16-13} focused on segmenting multi-word noun-compounds and assumed an is-a relation between the parts, as extracting \textit{(Francis Collins, is, NIH director)} from ``NIH director Francis Collins''. \newcite{XAVIER14.125} enriched the corpus with compound definitions from online dictionaries, for example, interpreting \textit{oil industry} as \textit{(industry, produces and delivers, oil)} based on the WordNet definition ``industry that produces and delivers oil''. This method is very limited as it can only interpret noun-compounds with dictionary entries, while the majority of English noun-compounds don't have them \cite{nakov2013interpretation}.

%% file: figures/model.tex
\resizebox{1\textwidth}{!}{
\begin{tabular}{|l|l|}
\hline
\input{figures/predict_w1} &
\input{figures/predict_p} \\ \hline
\end{tabular}
}

%% file: figures/predict_w1.tex
\begin{tikzpicture}[
  normal/.style={rectangle,draw,fill=white,minimum width=4cm,minimum height=0.7cm,rounded corners=.8ex},
  predicate_embedding/.style={rectangle,draw,fill=blue!20,minimum width=2cm,minimum height=0.6cm},
  special_embedding/.style={rectangle,draw,fill=my_orange!20,minimum width=2cm,minimum height=0.6cm},
  word_embedding/.style={rectangle,draw,fill=my_green!40,minimum width=3cm,    minimum height=0.6cm},
  every neuron/.style={circle,draw,minimum size=0.5cm},
  lstm_cell/.style={rectangle,draw,minimum width=1.5cm,minimum height=0.7cm},
  neuron missing/.style={draw=none,scale=2,execute at begin node=\color{black}$\cdots$}
  ]

    \node[word_embedding] (w1) {(23) made};
    \node[word_embedding] (w2) [below=0 of w1] {(28) apple};
    \node[word_embedding] (w3) [below=0 of w2] {(4145) cake};
    \node[word_embedding] (w4) [below=0 of w3] {...};
    \node[word_embedding] (w5) [below=0 of w4] {(7891) of};
    
    \node[special_embedding] (sw1) [right=1.25cm of w1] {(1) [w$_1$]};
    \node[special_embedding] (sw2) [below=0 of sw1] {(2) [w$_2$]};
    \node[special_embedding] (sp) [below=0 of sw2] {(3) [p]};
    
\node[word_embedding,rounded corners=.8ex,minimum width=2cm] (v_of) [above=0.7 of w1.north east] {of};
\node[word_embedding,rounded corners=.8ex,minimum width=2cm] (v_cake) [left=2.5 cm of v_of] {cake};
    \node[word_embedding,rounded corners=.8ex,minimum width=2cm] (v_made) [right=0.25 of v_cake] {made};
    \node[special_embedding,rounded corners=.8ex,minimum width=2cm] (v_w1) [right=0.25 of v_of] {[w$_1$]};    
     \path[draw,->] (w3.west) to [out=150,in=-90] (v_cake.south);
      \path[draw,->] (w1.west) to [out=150,in=-150] (v_made.south);
      \path[draw,->] (w5.east) to [out=30,in=-80] (v_of.south);
      \path[draw,->] (sw1.west) to [out=150,in=-150] (v_w1.south);

\node[lstm_cell] (lstm1) [above=0.2 of v_cake] {};
\node[lstm_cell] (lstm2) [above=0.2 of v_made] {};
\node[lstm_cell] (lstm3) [above=0.2 of v_of] {};
\node[lstm_cell] (lstm4) [above=0.2 of v_w1] {};
\path[draw,->] (v_cake.north) to (lstm1.south);
\path[draw,->] (v_made.north) to (lstm2.south);
\path[draw,->] (v_of.north) to (lstm3.south);
\path[draw,->] (v_w1.north) to (lstm4.south);
\path[draw,->] ($(lstm1.east)+(0,-0.1)$) to 
($(lstm2.west)+(0,-0.1)$);
\path[draw,->] ($(lstm2.west)+(0,0.1)$) to 
($(lstm1.east)+(0,0.1)$);
\path[draw,->] ($(lstm2.east)+(0,-0.1)$) to 
($(lstm3.west)+(0,-0.1)$);
\path[draw,->] ($(lstm3.west)+(0,0.1)$) to 
($(lstm2.east)+(0,0.1)$);
\path[draw,->] ($(lstm3.east)+(0,-0.1)$) to 
($(lstm4.west)+(0,-0.1)$);
\path[draw,->] ($(lstm4.west)+(0,0.1)$) to 
($(lstm3.east)+(0,0.1)$);

\node[normal,rounded corners=0ex,minimum height=1cm,minimum width=1.5cm] (mlpw) [above=0.5 cm of lstm4] {$MLP_w$};
\path[draw,->] (lstm4.north) to 
(mlpw.south);

\node (predicted_w1) [above=0.2 cm of mlpw] {$\hat{w_1}_i = 28$};
\path[draw,->] (mlpw) to (predicted_w1);
\end{tikzpicture}

%% file: figures/predict_p.tex
\begin{tikzpicture}[
  normal/.style={rectangle,draw,fill=white,minimum width=4cm,minimum height=0.7cm,rounded corners=.8ex},
  predicate_embedding/.style={rectangle,draw,fill=blue!20,minimum width=2cm,minimum height=0.6cm},
  special_embedding/.style={rectangle,draw,fill=my_orange!20,minimum width=2cm,minimum height=0.6cm},
  word_embedding/.style={rectangle,draw,fill=my_green!40,minimum width=3cm,    minimum height=0.6cm},
  every neuron/.style={circle,draw,minimum size=0.5cm},
  lstm_cell/.style={rectangle,draw,minimum width=1.5cm,minimum height=0.7cm},
  neuron missing/.style={draw=none,scale=2,execute at begin node=\color{black}$\cdots$}
  ]

    \node[word_embedding] (w1) {(23) made};
    \node[word_embedding] (w2) [below=0 of w1] {(28) apple};
    \node[word_embedding] (w3) [below=0 of w2] {(4145) cake};
    \node[word_embedding] (w4) [below=0 of w3] {...};
    \node[word_embedding] (w5) [below=0 of w4] {(7891) of};
    
    \node[special_embedding] (sw1) [right=1.25cm of w3] {(1) [w$_1$]};
    \node[special_embedding] (sw2) [below=0 of sw1] {(2) [w$_2$]};
    \node[special_embedding] (sp) [below=0 of sw2] {(3) [p]};
    
    \node[predicate_embedding] (p1) [right=3.5cm of w2,minimum width=4.5cm] {(78) [w$_2$] containing [w$_1$]};
    \node[predicate_embedding] (p2) [below=0 of p1,minimum width=4.5cm] {...};
    \node[predicate_embedding] (p3) [below=0 of p2,minimum width=4.5cm] {(131) [w$_2$] made of [w$_1$]};
    \node[predicate_embedding] (p4) [below=0 of p3,minimum width=4.5cm] {...};
    
  \node[special_embedding,rounded corners=.8ex,minimum width=2cm] (v_paraphrase) [above right=0.7 and 0.05 cm of w1.north east]{[p]};
\node[word_embedding,rounded corners=.8ex,minimum width=2cm] (v_cake) [left=0.25 cm of v_paraphrase] {cake};
    \node[word_embedding,rounded corners=.8ex,minimum width=2cm] (v_apple) [right=0.25 cm of v_paraphrase] {apple};
      \path[draw,->] (w2.east) to [out=0,in=-90] (v_apple.south);
      \path[draw,->] (w3.west) to [out=180,in=-180] (v_cake.west);
       \path[draw,->] (sp.west) to [out=180,in=-90] (v_paraphrase.south);

\node[lstm_cell] (lstm1) [above=0.2 of v_cake] {};
\node[lstm_cell] (lstm2) [above=0.2 of v_paraphrase] {};
\node[lstm_cell] (lstm3) [above=0.2 of v_apple] {};
\path[draw,->] (v_cake.north) to (lstm1.south);
\path[draw,->] (v_paraphrase.north) to (lstm2.south);
\path[draw,->] (v_apple.north) to (lstm3.south);
\path[draw,->] ($(lstm1.east)+(0,-0.1)$) to 
($(lstm2.west)+(0,-0.1)$);
\path[draw,->] ($(lstm2.west)+(0,0.1)$) to 
($(lstm1.east)+(0,0.1)$);
\path[draw,->] ($(lstm2.east)+(0,-0.1)$) to 
($(lstm3.west)+(0,-0.1)$);
\path[draw,->] ($(lstm3.west)+(0,0.1)$) to 
($(lstm2.east)+(0,0.1)$);

\node[normal,rounded corners=0ex,minimum height=1cm,minimum width=1.5cm] (mlpp) [above=0.5 cm of lstm2] {$MLP_p$};
\path[draw,->] (lstm2.north) to 
(mlpp.south);

\node (predicted_p) [above=0.2 cm of mlpp] {$\hat{p}_i = 78$};
\path[draw,->] (mlpp) to (predicted_p);

\end{tikzpicture}

%% file: model.tex
As opposed to previous approaches, that focus on predicting a paraphrase template for a given noun-compound, we reformulate the task as a multi-task learning problem (Section~\ref{sec:mtl}), and train the model to also predict a missing constituent given the paraphrase template and the other constituent. Our model is semi-supervised, and it expects as input a set of noun-compounds and a set of constrained part-of-speech tag-based templates that make valid prepositional and verbal paraphrases. Section~\ref{sec:wk_training_corpus} details the creation of training data, and Section~\ref{sec:wk_model_model} describes the model.

\subsection{Multi-task Reformulation} 
\label{sec:mtl}

Each training example consists of two constituents and a paraphrase $(w_2, p, w_1)$, and we train the model on 3 subtasks: (1) predict $p$ given $w_1$ and $w_2$, (2) predict $w_1$ given $p$ and $w_2$, and (3) predict $w_2$ given $p$ and $w_1$. Figure~\ref{fig:model} demonstrates the predictions for subtasks (1) (right) and (2) (left) for the training example (\textit{cake, made of, apple}). Effectively, the model is trained to answer questions such as ``what can cake be made of?'', ``what can be made of apple?'', and ``what are the possible relationships between cake and apple?''.

The multi-task reformulation helps learning better representations for paraphrase templates, by embedding semantically-similar paraphrases in proximity. Similarity between paraphrases stems either from lexical similarity and overlap between the paraphrases (e.g. `is made of' and `made of'), or from shared constituents, e.g. `[w$_2$] involved in [w$_1$]' and `[w$_2$] in [w$_1$] industry' can share [w$_1$] = \textit{insurance} and [w$_2$] = \textit{company}. This allows the model to predict a correct paraphrase for a given noun-compound, even when the constituents \emph{do not occur with that paraphrase in the corpus}. 

\subsection{Training Data} 
\label{sec:wk_training_corpus}

We collect a training set of $(w_2, p, w_1, s)$ examples, where $w_1$ and $w_2$ are constituents of a noun-compound $w_1 w_2$, $p$ is a templated paraphrase, and $s$ is the score assigned to the training instance.\footnote{We refer to ``paraphrases'' and ``paraphrase templates'' interchangeably. In the extracted templates, [w$_2$] always precedes [w$_1$], probably because w$_2$ is normally the head noun.}

We use the 19,491 noun-compounds found in the  SemEval tasks datasets \cite{butnariu-EtAl:2009:SEW,S13-2025} and in \newcite{tratz2011semantically}. To extract patterns of part-of-speech tags that can form noun-compound paraphrases, such as `[w$_2$] \textsc{verb} \textsc{prep} [w$_1$]', we use the SemEval task training data, but we do not use the lexical information in the gold paraphrases.

\paragraph{Corpus.} Similarly to previous noun-compound paraphrasing approaches, we use the Google N-gram corpus \cite{brants2006web} as a source of paraphrases \cite{wubben2010uvt,li2010ucd,surtani2013iiit,versley2013sfs}. The corpus consists of 
sequences of $n$ terms (for $n \in \{3,4,5\}$) that occur more than 40 times
on the web. We search for n-grams following the extracted patterns and containing $w_1$ and $w_2$'s lemmas for some noun-compound in the set. We remove punctuation, adjectives, adverbs and some determiners to unite similar paraphrases. For example, from the 5-gram `\textit{cake made of sweet apples}' we extract the training example \textit{(cake, made of, apple)}. We keep only paraphrases that occurred at least 5 times, resulting in 136,609 instances.

\paragraph{Weighting.} Each n-gram in the corpus is accompanied with its frequency, which we use to assign scores to the different paraphrases. For instance, `\textit{cake of apples}' may also appear in the corpus, although with lower frequency than `\textit{cake from apples}'. As also noted by \newcite{surtani2013iiit}, the shortcoming of such a weighting mechanism is that it prefers shorter paraphrases, which are much more common in the corpus (e.g. count(`\textit{cake made of apples}') $\ll$ count(`\textit{cake of apples}')). We overcome this by normalizing the frequencies \emph{for each paraphrase length}, creating a distribution of paraphrases in a given length.

\paragraph{Negative Samples.} We add 1\% of negative samples by selecting random corpus words $w_1$ and $w_2$ that do not co-occur, and adding an example ($w_2$, [w$_2$] is unrelated to [w$_1$], $w_1$, $s_n$), for some predefined negative samples score $s_n$. Similarly, for a word $w_i$ that did not occur in a paraphrase $p$ we add ($w_i$, $p$, UNK, $s_n$) or (UNK, $p$, $w_i$, $s_n$), where UNK is the unknown word. This may help the model deal with non-compositional noun-compounds, where $w_1$ and $w_2$ are unrelated, rather than forcibly predicting some relation between them. 

\subsection{Model}
\label{sec:wk_model_model}

For a training instance $(w_2, p, w_1, s)$, we predict each item given the encoding of the other two. 

\paragraph{Encoding.} We use the 100-dimensional pre-trained GloVe embeddings \cite{pennington-socher-manning:2014:EMNLP2014}, which are fixed during training. In addition, we learn embeddings for the special words [w$_1$], [w$_2$], and [p], which are used to represent a missing component, as in ``cake made of [w$_1$]'',  ``[w$_2$] made of apple'', and ``cake [p] apple''.

For a missing component $x \in $ \{[p], [w$_1$], [w$_2$]\} surrounded by the sequences of words $v_{1:i-1}$ and $v_{i+1:n}$, we encode the sequence using a bidirectional long-short term memory (bi-LSTM) network \cite{graves2005framewise}, and take the $i$th output vector as representing the missing component: $bLS(v_{1:i}, x, v_{i+1:n})_i$.

In bi-LSTMs, each output vector is a concatenation of the outputs of the forward and backward LSTMs, so the output vector is expected to contain information on valid substitutions both with respect to the previous words $v_{1:i-1}$ and the subsequent words $v_{i+1:n}$.

\begin{table*}[t]
\scriptsize
\center
\input{figures/model_output_examples}
\vspace{-8pt}
\caption{Examples of top ranked predicted components using the model: predicting the paraphrase given $w_1$ and $w_2$ (left), $w_1$ given $w_2$ and the paraphrase (middle), and $w_2$ given $w_1$ and the paraphrase (right).}
\label{tab:model_output_examples}
\vspace{-12pt}
\end{table*}

\begin{figure*}[t]
\center
\includegraphics[width=\textwidth]{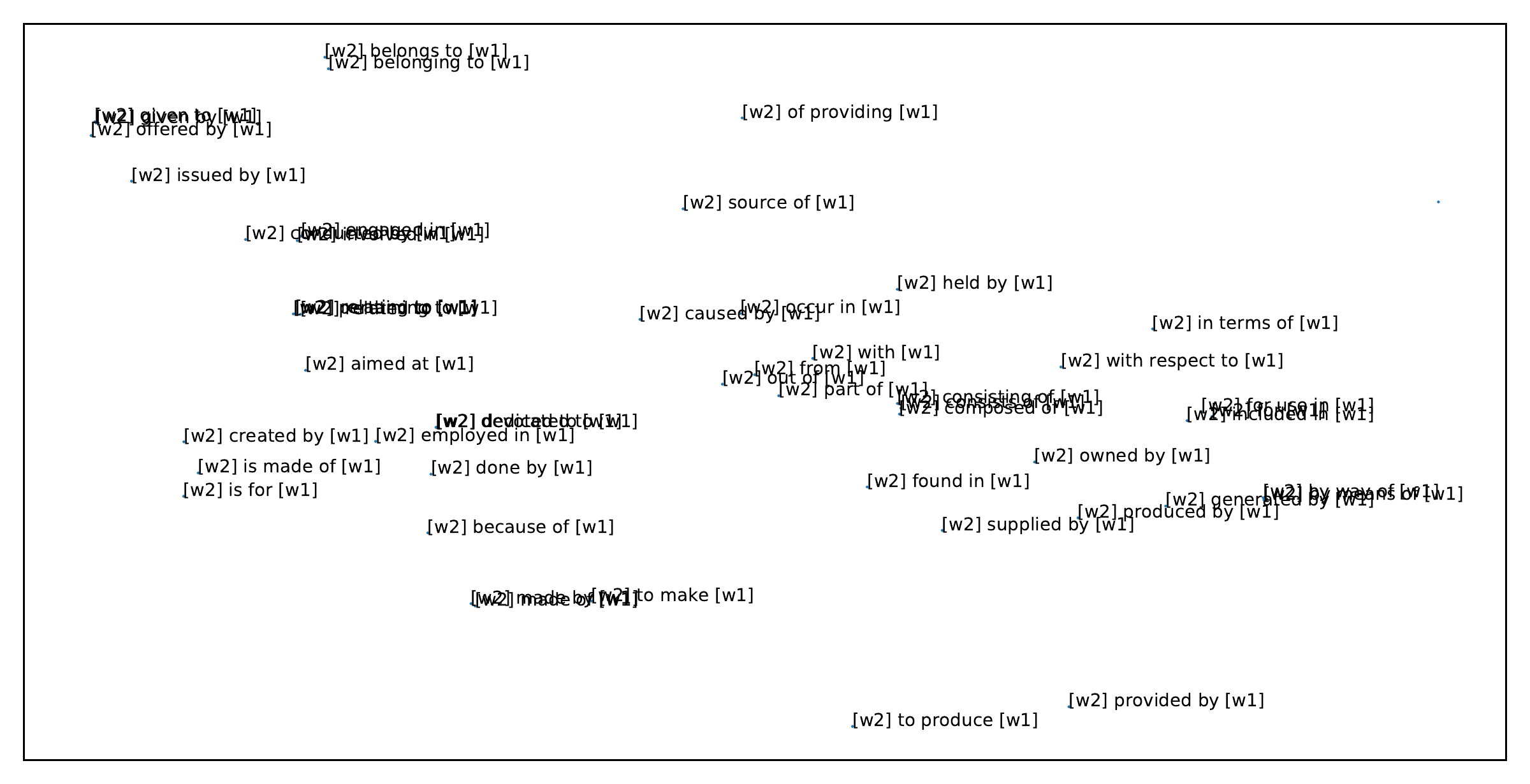}
\vspace{-32pt}
\caption{A t-SNE map of a sample of paraphrases, using the paraphrase vectors encoded by the biLSTM, for example $bLS($[w$_2$] made of [w$_1$]$)$.}
\label{fig:paraphrase_tsne}
\vspace{-5pt}
\end{figure*}

\paragraph{Prediction.} We predict a distribution of the vocabulary of the missing component, i.e. to predict $w_1$ correctly we need to predict its index in the word vocabulary $V_w$, while the prediction of $p$ is from the vocabulary of paraphrases in the training set, $V_p$. We predict the following distributions:
\vspace{-15pt}

\begin{equation}
\hspace{-2pt}
\begin{aligned}
       & \hat{p} =  \operatorname{softmax} (W_p \cdot bLS(\vec{w_2}, [p], \vec{w_1})_2) \\
      & \hat{w_1} =  \operatorname{softmax} (W_w \cdot bLS(\vec{w_2}, \vec{p}_{1:n}, [w_1])_{n+1}) \\ 
      & \hat{w_2} =  \operatorname{softmax} (W_w \cdot bLS([w_2], \vec{p}_{1:n}, \vec{w_1})_1) \\
\end{aligned}
\label{eq:distributions}
\hspace{-9pt}
\end{equation}

\vspace{-7pt}
\noindent where $W_w \in \mathcal{R}^{|V_w| \times 2d}$, $W_p \in \mathcal{R}^{|V_p| \times 2d}$, and $d$ is the embeddings dimension. 

During training, we compute cross-entropy loss for each subtask using the gold item and the prediction, sum up the losses, and weight them by the instance score. During inference, we predict the missing components by picking the best scoring index in each distribution:\footnote{In practice, we pick the $k$ best scoring indices in each distribution for some predefined $k$, as we discuss in Section~\ref{sec:tasks}.}
\vspace{-12pt}

\begin{equation}
\begin{aligned}
        \hat{p}_i = \operatorname{argmax} (\hat{p}) \\
        \hat{w_1}_i = \operatorname{argmax} (\hat{w_1}) \\
        \hat{w_2}_i = \operatorname{argmax} (\hat{w_2})
\end{aligned}
\end{equation}
\vspace{-12pt}

The subtasks share the pre-trained word embeddings, the special embeddings, and the biLSTM parameters. Subtasks (2) and (3) also share $W_w$, the MLP that predicts the index of a word. 

\paragraph{Implementation Details.} The model is implemented in DyNet \cite{neubig2017dynet}. We dedicate a small number of noun-compounds from the corpus for validation. We train for up to 10 epochs, stopping early if the validation loss has not improved in 3 epochs. We use Momentum SGD \cite{nesterov1983method}, and set the batch size to 10 and the other hyper-parameters to their default values. 

%% file: figures/model_output_examples.tex
\hspace*{-20pt}
\begin{tabular}{ c c c || c c c || c c c  }
    \toprule
    {[}w$_1$] & {[}w$_2$] & \textbf{Predicted Paraphrases} & {[}w$_2$] & \textbf{Paraphrase} & \textbf{Predicted} [w$_1$] &  \textbf{Paraphrase} & [w$_1$] & \textbf{Predicted} [w$_2$] \\ 				
	\hline
        \multirow{4}{*}{cataract} &  \multirow{4}{*}{surgery} & {[}w$_2$] of [w$_1$] & \multirow{4}{*}{surgery} & \multirow{4}{*}{{[}w$_2$] to treat [w$_1$]} & heart & \multirow{4}{*}{{[}w$_2$] to treat [w$_1$]} & \multirow{4}{*}{cataract} & surgery \\
        & & {[}w$_2$] on [w$_1$] & & & brain & & & drug \\
		& & {[}w$_2$] to remove [w$_1$] & & & back & & & patient \\
		& & {[}w$_2$] in patients with [w$_1$] & & & knee & & & transplant \\
        \hline
		\multirow{4}{*}{software} & \multirow{4}{*}{company} & {[}w$_2$] of [w$_1$] & \multirow{4}{*}{company} & \multirow{4}{*}{{[}w$_2$] engaged in [w$_1$]}	& management & \multirow{4}{*}{{[}w$_2$] engaged in [w$_1$]} & \multirow{4}{*}{software} & company \\
		& & {[}w$_2$] to develop [w$_1$] & & & production & & & firm \\
		& & {[}w$_2$] in [w$_1$] industry & & & computer & & & engineer \\
        & & {[}w$_2$] involved in [w$_1$] & & & business & & & industry \\
        \hline
	\multirow{4}{*}{stone} & \multirow{4}{*}{wall}  & {[}w$_2$] is of [w$_1$] & \multirow{4}{*}{meeting} & \multirow{4}{*}{{[}w$_2$] held in [w$_1$]} & spring & \multirow{4}{*}{{[}w$_2$] held in [w$_1$]} & \multirow{4}{*}{morning} & party \\
	& & {[}w$_2$] of [w$_1$] & & & afternoon & & & meeting \\
	& & {[}w$_2$] is made of [w$_1$] & & & hour & & & rally \\
	& & {[}w$_2$] made of [w$_1$] & & & day & & & session \\
		\bottomrule
	\end{tabular}
	

%% file: qualitative_analysis.tex
To estimate the quality of the proposed model, we first provide a qualitative analysis of the model outputs. Table~\ref{tab:model_output_examples} displays examples of the model outputs for each possible usage: predicting the paraphrase given the constituent words, and predicting each constituent word given the paraphrase and the other word. 

The examples in the table are from among the top 10 ranked predictions for each component-pair. We note that most of the ($w_2$, paraphrase, $w_1$) triplets in the table do not occur in the training data, but are rather generalized from similar examples. For example, there is no training instance for ``company in the software industry'' but there is a ``firm in the software industry'' and a company in many other industries. 

While the frequent prepositional paraphrases are often ranked at the top of the list, the model also retrieves more specified verbal paraphrases. The list often contains multiple semantically-similar paraphrases, such as `[w$_2$] involved in [w$_1$]' and `[w$_2$] in [w$_1$] industry'. This is a result of the model training objective (Section~\ref{sec:model}) which positions the vectors of semantically-similar paraphrases close to each other in the embedding space, based on similar constituents.  

To illustrate paraphrase similarity we compute a t-SNE projection \cite{van2014accelerating} of the embeddings of all the paraphrases, and draw a sample of 50 paraphrases in Figure~\ref{fig:paraphrase_tsne}. The projection positions semantically-similar but lexically-divergent paraphrases in proximity, likely due to many shared constituents. For instance, `with', `from', and `out of' can all describe the relation between food words and their ingredients.

%% file: paraphrasing.tex
\subsection{Paraphrasing}
\label{sec:main_paraphrasing}

\begin{table*}[t]
\small
\center
\input{figures/paraphrasing_results}
\vspace{-8pt}
\caption{Results of the proposed method and the baselines on the  SemEval 2013 task.}
\label{tab:paraphrasing_results}
\vspace{-10pt}
\end{table*}

\paragraph{Task Definition.} The general goal of this task is to interpret each noun-compound to multiple prepositional and verbal paraphrases. In SemEval 2013 Task 4,\footnote{\scriptsize\url{https://www.cs.york.ac.uk/semeval-2013/task4}} the participating systems were asked to retrieve a ranked list of paraphrases for each noun-compound, which was automatically evaluated against a similarly ranked list of paraphrases proposed by human annotators.

\paragraph{Model.} For a given noun-compound $w_1 w_2$, we first predict the $k=250$ most likely paraphrases: $\hat{p}_1, ..., \hat{p}_k = \operatorname{argmax}_k \hat{p}$, where $\hat{p}$ is the distribution of paraphrases defined in Equation~\ref{eq:distributions}. 

While the model also provides a score for each paraphrase (Equation~\ref{eq:distributions}), the scores have not been optimized to correlate with human judgments. We therefore developed a re-ranking model that receives a list of paraphrases and re-ranks the list to better fit the human judgments.

We follow \newcite{herbrich2000large} and learn a pairwise ranking model. The model determines which of two paraphrases of the same noun-compound should be ranked higher, and it is implemented as an SVM classifier using scikit-learn \cite{scikit-learn}. For training, we use the available training data with gold paraphrases and ranks provided by the SemEval task organizers. We extract the following features for a paraphrase $p$:

\vspace{-8pt}
\begin{enumerate}[noitemsep,leftmargin=*] 
	\setlength{\itemsep}{1pt}
	\setlength{\parskip}{0pt}
	\setlength{\parsep}{0pt}
  \item The part-of-speech tags contained in $p$
  \item The prepositions contained in $p$
  \item The number of words in $p$
  \item Whether $p$ ends with the special [w$_1$] symbol
  \item $cosine(bLS([w_2], p, [w_1])_2, \vec{V_p}^{\hat{p}_i}) \cdot \hat{p}^{\hat{p}_i}$
\end{enumerate}
\vspace{-3pt}

\noindent where $\vec{V_p}^{\hat{p}_i}$ is the biLSTM encoding of the predicted paraphrase computed in Equation~\ref{eq:distributions} and $\hat{p}^{\hat{p}_i}$ is its confidence score. The last feature incorporates the original model score into the decision, as to not let other considerations such as preposition frequency  in the training set take over.

During inference, the model sorts the list of paraphrases retrieved for each noun-compound according to the pairwise ranking. It then scores each paraphrase by multiplying its rank with its original model score, and prunes paraphrases with final score $< 0.025$. The values for $k$ and the threshold were tuned on the training set.

\paragraph{Evaluation Settings.} The SemEval 2013 task provided a scorer that compares words and n-grams from the gold paraphrases against those in the predicted paraphrases, where agreement on a prefix of a word (e.g. in derivations) yields a partial scoring. The overall score assigned to each system is calculated in two different ways. The `isomorphic' setting rewards both precision and recall, and performing well on it requires accurately reproducing as many of the gold paraphrases as possible, and in much the same order. The `non-isomorphic' setting rewards only precision, and performing well on it requires accurately reproducing the top-ranked gold paraphrases, with no importance to order.

\paragraph{Baselines.} We compare our method with the published results from the SemEval task. The SemEval 2013 baseline generates for each noun-compound a list of prepositional paraphrases in an arbitrary fixed order. It achieves a moderately good score in the non-isomorphic setting by generating a fixed set of paraphrases which are both common and generic. The MELODI system performs similarly: it represents each noun-compound using a compositional distributional vector \cite{mitchell2010composition} which is then used to predict paraphrases from the corpus. The performance of MELODI indicates that the system was rather conservative, yielding a few common paraphrases rather than many specific ones. SFS and IIITH, on the other hand, show a more balanced trade-off between recall and precision.

As a sanity check, we also report the results of a baseline that retrieves ranked paraphrases from the training data collected in Section~\ref{sec:wk_training_corpus}. This baseline has no generalization abilities, therefore it is expected to score poorly on the recall-aware isomorphic setting.

\paragraph{Results.} Table~\ref{tab:paraphrasing_results} displays the performance of the proposed method and the baselines in the two evaluation settings. Our method outperforms all the methods in the isomorphic setting. In the non-isomorphic setting, it outperforms the other two systems that score reasonably on the isomorphic setting (SFS and IIITH) but cannot compete with the systems that focus on achieving high precision. 

The main advantage of our proposed model is in its ability to generalize, and that is also demonstrated in comparison to our baseline performance. The baseline retrieved paraphrases only for a third of the noun-compounds (61/181), expectedly yielding poor performance on the isomorphic setting. Our model, which was trained on the very same data, retrieved paraphrases for all noun-compounds. For example, \textit{welfare system} was not present in the training data, yet the model predicted the correct paraphrases ``system of welfare benefits'', ``system to provide welfare'' and others.

\begin{table}[!t]
	\center
	\small	\input{figures/error_analysis_paraphrasing}
	\vspace{-7pt}
	\caption{Categories of false positive and false negative predictions along with their percentage.}
	\label{tab:paraphrasing_error_analysis}
	\vspace{-12pt}
\end{table}

\paragraph{Error Analysis.} We analyze the causes of the false positive and false negative errors made by the model. For each error type we sample 10 noun-compounds. For each noun-compound, false positive errors are the top 10 predicted paraphrases which are not included in the gold paraphrases, while false negative errors are the top 10 gold paraphrases not found in the top $k$ predictions made by the model. Table~\ref{tab:paraphrasing_error_analysis} displays the manually annotated categories for each error type. 

Many false positive errors are actually valid paraphrases that were not suggested by the human annotators (error 1, ``discussion by group''). Some are borderline valid with minor grammatical changes (error 6, ``force of coalition forces'') or too specific (error 2, ``life \textit{of women in} community'' instead of ``life \textit{in} community''). Common prepositional paraphrases were often retrieved although they are incorrect (error 3). We conjecture that this error often stem from an n-gram that does not respect the syntactic structure of the sentence, e.g. a sentence such as ``rinse away the oil from baby 's head'' produces the n-gram ``oil from baby''. 

With respect to false negative examples, they consisted of many long paraphrases, while our model was restricted to 5 words due to the source of the training data (error 1, ``holding done in the case of a share''). Many prepositional paraphrases consisted of determiners, which we conflated with the same paraphrases without determiners (error 2, ``mutation of a gene''). Finally, in some paraphrases, the constituents in the gold paraphrase appear in inflectional forms (error 3, ``holding of shares'' instead of ``holding of share'').

%% file: figures/paraphrasing_results.tex
\begin{tabular}{c c c c }
\toprule
& Method & isomorphic & non-isomorphic \\ \hline
\multirow{4}{*}{Baselines} & SFS \cite{versley2013sfs} & 23.1 & 17.9 \\
& IIITH \cite{surtani2013iiit} & 23.1 & 25.8 \\
& MELODI \cite{vandecruys-afantenos-muller:2013:SemEval-20132} & 13.0 & 54.8 \\ 
& SemEval 2013 Baseline \cite{S13-2025} & 13.8 & 40.6 \\ 
\midrule
\multirow{2}{*}{This paper} & Baseline & 3.8 & 16.1 \\ 
& Our method & \textbf{28.2} & 28.4 \\ 
\bottomrule
\end{tabular}

%% file: figures/error_analysis_paraphrasing.tex
\begin{tabular}{ l  l }
    \toprule
    \textbf{Category} & \textbf{\%}  \\ 
	\midrule
	 \multicolumn{2}{l}{\textbf{False Positive}}  \\ 
        (1) Valid paraphrase missing from gold & 44\% \\
		(2) Valid paraphrase, slightly too specific & 15\% \\
		(3) Incorrect, common prepositional paraphrase & 14\% \\
		(4) Incorrect, other errors & 14\% \\
		(5) Syntactic error in paraphrase & 8\% \\
		(6) Valid paraphrase, but borderline grammatical & 5\% \\
	\midrule
	 \multicolumn{2}{l}{\textbf{False Negative}}  \\ 
		(1) Long paraphrase (more than 5 words) & 30\% \\
		(2) Prepositional paraphrase with determiners & 25\% \\
		(3) Inflected constituents in gold & 10\% \\
		(4) Other errors & 35\% \\
	\bottomrule
	\end{tabular}

%% file: classification.tex
\subsection{Classification}
\label{sec:main_classification}

Noun-compound classification is defined as a multiclass classification problem: given a pre-defined set of relations, classify $w_1 w_2$ to the relation that holds between $w_1$ and $w_2$. Potentially, the corpus co-occurrences of $w_1$ and $w_2$ may contribute to the classification, e.g. `[w$_2$] held at [w$_1$]' indicates a \smallsc{time} relation. \newcite{tratz-hovy:2010:ACL} included such features in their classifier, but ablation tests showed that these features had a relatively small contribution, probably due to the sparseness of the paraphrases. Recently, \newcite{nc_naacl2018} showed that paraphrases may contribute to the classification  when represented in a continuous space. 

\paragraph{Model.} We generate a paraphrase vector representation $\vec{par}(w_1 w_2)$ for a given noun-compound $w_1 w_2$ as follows. We predict the indices of the $k$ most likely paraphrases: $\hat{p}_1, ..., \hat{p}_k = \operatorname{argmax}_k \hat{p}$, where $\hat{p}$ is the distribution on the paraphrase vocabulary $V_p$, as defined in Equation~\ref{eq:distributions}. We then encode each paraphrase using the biLSTM, and average the paraphrase vectors, weighted by their confidence scores in $\hat{p}$: 

\vspace{-10pt}
\begin{equation}
\vec{par}(w_1 w_2) =  
\frac{\sum_{i=1}^k \hat{p}^{\hat{p}_i} \cdot \vec{V_p}^{\hat{p}_i}}{\sum_{i=1}^k \hat{p}^{\hat{p}_i}}
\label{eq:nc_par_representation}
\end{equation}

We train a linear classifier, and represent $w_1 w_2$ in a feature vector $f(w_1w_2)$ in two variants: \smallsf{paraphrase}: $f(w_1w_2) = \vec{par}(w_1 w_2)$, or \smallsf{integrated}: concatenated to the constituent word embeddings $f(w_1w_2) = [\vec{par}(w_1 w_2), \vec{w_1}, \vec{w_2}]$. The classifier type (logistic regression/SVM), $k$, and the penalty are tuned on the validation set. We also provide a baseline in which we ablate the paraphrase component from our model, representing a noun-compound by the concatenation of its constituent embeddings $f(w_1w_2) = [\vec{w_1}, \vec{w_2}]$ (\smallsf{distributional}). 

\paragraph{Datasets.} We evaluate on the \newcite{tratz2011semantically} dataset, which consists of 19,158 instances, labeled in 37 fine-grained relations (\smalltt{Tratz-fine}) or 12 coarse-grained relations (\smalltt{Tratz-coarse}). 

We report the performance on two different dataset splits to train, test, and validation: a random split in a 75:20:5 ratio, and, following concerns raised by \newcite{W16-1604} about lexical memorization \cite{levy-EtAl:2015:NAACL-HLT}, on a lexical split in which the sets consist of distinct vocabularies. The lexical split better demonstrates the scenario in which a noun-compound whose constituents have not been observed needs to be interpreted based on similar observed noun-compounds, e.g. inferring the relation in \textit{pear tart} based on \textit{apple cake} and other similar compounds. We follow the random and full-lexical splits from \newcite{nc_naacl2018}. 

\paragraph{Baselines.} We report the results of 3 baselines representative of different approaches: 

\noindent 1) \textbf{Feature-based} \cite{tratz-hovy:2010:ACL}: we re-implement a version of the classifier with features from WordNet and Roget's Thesaurus.\\
\noindent 2) \textbf{Compositional} \cite{W16-1604}: a neural architecture that operates on the distributional representations of the noun-compound and its constituents. Noun-compound representations are learned with the Full-Additive \cite{zanzotto2010estimating} and Matrix \cite{D12-1110} models. We report the results from \newcite{nc_naacl2018}.\\
\noindent 3) \textbf{Paraphrase-based} \cite{nc_naacl2018}: a neural classification model that learns an LSTM-based representation of the joint occurrences of $w_1$ and $w_2$ in a corpus (i.e. observed paraphrases), and integrates distributional information using the constituent embeddings. 

\begin{table}[t]
\small
\center
\input{figures/classification_results}
\vspace{-10pt}
\caption{Classification results. For each dataset split, the top part consists of baseline methods and the bottom part of methods from this paper. The best performance in each part appears in bold.}
\label{tab:classification_results}
\vspace{-12pt}
\end{table}

\paragraph{Results.} Table~\ref{tab:classification_results} displays the methods' performance on the two versions of the \newcite{tratz2011semantically} dataset and the two dataset splits. The \smallsf{paraphrase} model on its own is inferior to the \smallsf{distributional} model, however, the \smallsf{integrated} version improves upon the \smallsf{distributional} model in 3 out of 4 settings, demonstrating the complementary nature of the distributional and paraphrase-based methods. The contribution of the paraphrase component is especially noticeable in the lexical splits. 

As expected, the integrated method in \newcite{nc_naacl2018}, in which the paraphrase representation was trained with the objective of classification, performs better than our integrated model. The superiority of both integrated models in the lexical splits confirms that paraphrases are beneficial for classification. 

\begin{table*}[!t]
	\center
	\small	\input{figures/classification_integrated_corrected}
	\vspace{-7pt}
	\caption{Examples of noun-compounds that were correctly classified by the \smallsf{integrated} model while being incorrectly classified by \smallsf{distributional}, along with top ranked indicative paraphrases.}
	\label{tab:qualitative}
	\vspace{-12pt}
\end{table*}

\paragraph{Analysis.} To analyze the contribution of the paraphrase component to the classification, we focused on the differences between the \smallsf{distributional} and \smallsf{integrated} models on the \smalltt{Tratz-Coarse} lexical split. Examination of the per-relation $F_1$ scores revealed that the relations for which performance improved the most in the \smallsf{integrated} model were \smallsc{topical} (+11.1 $F_1$ points), \smallsc{objective} (+5.5), \smallsc{attribute} (+3.8) and \smallsc{location/part\_whole} (+3.5).

Table~\ref{tab:qualitative} provides examples of noun-compounds that were correctly classified by the \smallsf{integrated} model while being incorrectly classified by the \smallsf{distributional} model. For each noun-compound, we provide examples of top ranked paraphrases which are indicative of the gold label relation. 

%% file: figures/classification_results.tex
\begin{tabular}{ c | c c }
\toprule
\textbf{Dataset \& Split} & \textbf{Method} & $\mathbf{F_1}$ \\ 
\hline
\multirow{6}{*}{\specialcell{\texttt{Tratz}\\\texttt{fine}\\Random}} & \newcite{tratz-hovy:2010:ACL} & \textbf{0.739} \\
& \newcite{W16-1604} & 0.725 \\
& \newcite{nc_naacl2018} & 0.714 \\
\hhline{~--}
& \textsf{distributional} & \textbf{0.677} \\
& \textsf{paraphrase} & 0.505 \\ 
& \textsf{integrated} & 0.673 \\
\hline
\multirow{6}{*}{\specialcell{\texttt{Tratz}\\\texttt{fine}\\Lexical}} & \newcite{tratz-hovy:2010:ACL} & 0.340 \\
& \newcite{W16-1604} & 0.334 \\
& \newcite{nc_naacl2018} & \textbf{0.429} \\
\hhline{~--}
& \textsf{distributional} & 0.356 \\
& \textsf{paraphrase} & 0.333 \\ 
& \textsf{integrated} & \textbf{0.370} \\
\hline
\multirow{6}{*}{\specialcell{\texttt{Tratz}\\\texttt{coarse}\\Random}} & \newcite{tratz-hovy:2010:ACL} & 0.760 \\
& \newcite{W16-1604} & \textbf{0.775} \\
& \newcite{nc_naacl2018} & 0.736 \\
\hhline{~--}
& \textsf{distributional} & 0.689 \\
& \textsf{paraphrase} & 0.557 \\ 
& \textsf{integrated} & \textbf{0.700} \\
\hline
\multirow{6}{*}{\specialcell{\texttt{Tratz}\\\texttt{coarse}\\Lexical}} & \newcite{tratz-hovy:2010:ACL} & 0.391 \\
& \newcite{W16-1604} & 0.372 \\
& \newcite{nc_naacl2018} & \textbf{0.478} \\
\hhline{~--}
& \textsf{distributional} & 0.370 \\
& \textsf{paraphrase} & 0.345 \\ 
& \textsf{integrated} & \textbf{0.393} \\
\bottomrule
\end{tabular}

%% file: figures/classification_integrated_corrected.tex
\hspace*{-15pt}
\begin{tabular}{ c  c  c  c }
    \toprule
    \textbf{Example Noun-compounds} & \textbf{Gold} & \textbf{Distributional} & \textbf{Example Paraphrases} \\ 
	\midrule
        \textit{printing plant} & \smallsc{purpose} & \smallsc{objective} &  [w$_2$] engaged in [w$_1$] \\
        \midrule
\specialcell{\textit{marketing expert} \\ \textit{development expert}} & \smallsc{topical} & \smallsc{objective} & \specialcell{[w$_2$] in [w$_1$] \\ {[}w$_2$] knowledge of [w$_1$]} \\
\midrule
\textit{weight/job loss} & \smallsc{objective} & \smallsc{causal} & [w$_2$] of [w$_1$] \\ 
\midrule
		\specialcell{\textit{rubber band}\\ \textit{rice cake}} & \smallsc{containment} & \smallsc{purpose} & \specialcell{[w$_2$] made of [w$_1$] \\ {[}w$_2$] is made of [w$_1$]} \\
        \midrule
		\textit{laboratory animal} & \smallsc{location/part-whole} & \smallsc{attribute} & [w$_2$] in [w$_1$], [w$_2$] used in [w$_1$] \\		
	\bottomrule
	\end{tabular}

%% file: compositionality.tex
Our paraphrasing approach at its core assumes compositionality: only a noun-compound whose meaning is derived from the meanings of its constituent words can be rephrased using them. In \S\ref{sec:wk_training_corpus} we added negative samples to the training data to simulate non-compositional noun-compounds, which are included in the classification dataset (\S\ref{sec:main_classification}). We assumed that these compounds, more often than compositional ones would consist of unrelated constituents (\textit{spelling bee}, \textit{sacred cow}), and added instances of random unrelated nouns with `[w$_2$] is unrelated to [w$_1$]'. Here, we assess whether our model succeeds to recognize non-compositional noun-compounds.

We used the compositionality dataset of \newcite{reddy-mccarthy-manandhar:2011:IJCNLP-2011} which consists of 90 noun-compounds along with human judgments about their compositionality in a scale of 0-5, 0 being non-compositional and 5 being compositional. For each noun-compound in the dataset, we predicted the 15 best paraphrases and analyzed the errors. The most common error was predicting paraphrases for idiomatic compounds which may have a plausible concrete interpretation or which originated from one. For example, it predicted that \textit{silver spoon} is simply a \textit{spoon} made of \textit{silver} and that \textit{monkey business} is a \textit{business} that buys or raises \textit{monkeys}. In other cases, it seems that the strong prior on one constituent leads to ignoring the other, unrelated constituent, as in predicting ``\textit{wedding} made of \textit{diamond}''. Finally, the ``unrelated'' paraphrase was predicted for a few compounds, but those are not necessarily non-compositional (\textit{application form}, \textit{head teacher}). We conclude that the model does not address compositionality and suggest to apply it only to compositional compounds, which may be recognized using compositionality prediction methods as in \newcite{reddy-mccarthy-manandhar:2011:IJCNLP-2011}.

%% file: main.bbl
\begin{thebibliography}{}
\expandafter\ifx\csname natexlab\endcsname\relax\def\natexlab#1{#1}\fi

\bibitem[{Barzilay and McKeown(2001)}]{P01-1008}
Regina Barzilay and R.~Kathleen McKeown. 2001.
\newblock \href{http://aclweb.org/anthology/P01-1008}{Extracting paraphrases
  from a parallel corpus}.
\newblock In {\em Proceedings of the 39th Annual Meeting of the Association for
  Computational Linguistics\/}.
\newblock
  \href{http://aclweb.org/anthology/P01-1008}{http://aclweb.org/anthology/P01-1008}.

\bibitem[{Brants and Franz(2006)}]{brants2006web}
Thorsten Brants and Alex Franz. 2006.
\newblock Web 1t 5-gram version 1 .

\bibitem[{Butnariu et~al.(2009)Butnariu, Kim, Nakov, \'{O}~S\'{e}aghdha,
  Szpakowicz, and Veale}]{butnariu-EtAl:2009:SEW}
Cristina Butnariu, Su~Nam Kim, Preslav Nakov, Diarmuid \'{O}~S\'{e}aghdha, Stan
  Szpakowicz, and Tony Veale. 2009.
\newblock \href{http://www.aclweb.org/anthology/W09-2416}{Semeval-2010 task 9:
  The interpretation of noun compounds using paraphrasing verbs and
  prepositions}.
\newblock In {\em Proceedings of the Workshop on Semantic Evaluations: Recent
  Achievements and Future Directions (SEW-2009)\/}. Association for
  Computational Linguistics, Boulder, Colorado, pages 100--105.
\newblock
  \href{http://www.aclweb.org/anthology/W09-2416}{http://www.aclweb.org/anthology/W09-2416}.

\bibitem[{Dima(2016)}]{W16-1604}
Corina Dima. 2016.
\newblock {\em Proceedings of the 1st Workshop on Representation Learning for
  NLP\/}, Association for Computational Linguistics, chapter On the
  Compositionality and Semantic Interpretation of English Noun Compounds, pages
  27--39.
\newblock
  \href{https://doi.org/10.18653/v1/W16-1604}{https://doi.org/10.18653/v1/W16-1604}.

\bibitem[{Dima and Hinrichs(2015)}]{dima2015automatic}
Corina Dima and Erhard Hinrichs. 2015.
\newblock Automatic noun compound interpretation using deep neural networks and
  word embeddings.
\newblock {\em IWCS 2015\/} page 173.

\bibitem[{Etzioni et~al.(2008)Etzioni, Banko, Soderland, and
  Weld}]{etzioni2008open}
Oren Etzioni, Michele Banko, Stephen Soderland, and Daniel~S Weld. 2008.
\newblock Open information extraction from the web.
\newblock {\em Communications of the ACM\/} 51(12):68--74.

\bibitem[{Ganitkevitch et~al.(2013)Ganitkevitch, Van~Durme, and
  Callison-Burch}]{N13-1092}
Juri Ganitkevitch, Benjamin Van~Durme, and Chris Callison-Burch. 2013.
\newblock \href{http://aclweb.org/anthology/N13-1092}{P{P}{D}{B}: The
  paraphrase database}.
\newblock In {\em Proceedings of the 2013 Conference of the North American
  Chapter of the Association for Computational Linguistics: Human Language
  Technologies\/}. Association for Computational Linguistics, pages 758--764.
\newblock
  \href{http://aclweb.org/anthology/N13-1092}{http://aclweb.org/anthology/N13-1092}.

\bibitem[{Girju(2007)}]{girju:2007:ACLMain}
Roxana Girju. 2007.
\newblock \href{http://www.aclweb.org/anthology/P07-1072}{Improving the
  interpretation of noun phrases with cross-linguistic information}.
\newblock In {\em Proceedings of the 45th Annual Meeting of the Association of
  Computational Linguistics\/}. Association for Computational Linguistics,
  Prague, Czech Republic, pages 568--575.
\newblock
  \href{http://www.aclweb.org/anthology/P07-1072}{http://www.aclweb.org/anthology/P07-1072}.

\bibitem[{Graves and Schmidhuber(2005)}]{graves2005framewise}
Alex Graves and J{\"u}rgen Schmidhuber. 2005.
\newblock Framewise phoneme classification with bidirectional lstm and other
  neural network architectures.
\newblock {\em Neural Networks\/} 18(5-6):602--610.

\bibitem[{Hendrickx et~al.(2013)Hendrickx, Kozareva, Nakov, {\'O}~S{\'e}aghdha,
  Szpakowicz, and Veale}]{S13-2025}
Iris Hendrickx, Zornitsa Kozareva, Preslav Nakov, Diarmuid {\'O}~S{\'e}aghdha,
  Stan Szpakowicz, and Tony Veale. 2013.
\newblock \href{http://aclweb.org/anthology/S13-2025}{Semeval-2013 task 4: Free
  paraphrases of noun compounds}.
\newblock In {\em Second Joint Conference on Lexical and Computational
  Semantics (*SEM), Volume 2: Proceedings of the Seventh International Workshop
  on Semantic Evaluation (SemEval 2013)\/}. Association for Computational
  Linguistics, pages 138--143.
\newblock
  \href{http://aclweb.org/anthology/S13-2025}{http://aclweb.org/anthology/S13-2025}.

\bibitem[{Herbrich(2000)}]{herbrich2000large}
Ralf Herbrich. 2000.
\newblock Large margin rank boundaries for ordinal regression.
\newblock {\em Advances in large margin classifiers\/} pages 115--132.

\bibitem[{Kim and Nakov(2011)}]{D11-1060}
Nam~Su Kim and Preslav Nakov. 2011.
\newblock \href{http://aclweb.org/anthology/D11-1060}{Large-scale noun compound
  interpretation using bootstrapping and the web as a corpus}.
\newblock In {\em Proceedings of the 2011 Conference on Empirical Methods in
  Natural Language Processing\/}. Association for Computational Linguistics,
  pages 648--658.
\newblock
  \href{http://aclweb.org/anthology/D11-1060}{http://aclweb.org/anthology/D11-1060}.

\bibitem[{Kim and Baldwin(2007)}]{kim2007interpreting}
Su~Nam Kim and Timothy Baldwin. 2007.
\newblock Interpreting noun compounds using bootstrapping and sense
  collocation.
\newblock In {\em Proceedings of Conference of the Pacific Association for
  Computational Linguistics\/}. pages 129--136.

\bibitem[{Levy et~al.(2015)Levy, Remus, Biemann, and
  Dagan}]{levy-EtAl:2015:NAACL-HLT}
Omer Levy, Steffen Remus, Chris Biemann, and Ido Dagan. 2015.
\newblock \href{http://www.aclweb.org/anthology/N15-1098}{Do supervised
  distributional methods really learn lexical inference relations?}
\newblock In {\em Proceedings of the 2015 Conference of the North American
  Chapter of the Association for Computational Linguistics: Human Language
  Technologies\/}. Association for Computational Linguistics, Denver, Colorado,
  pages 970--976.
\newblock
  \href{http://www.aclweb.org/anthology/N15-1098}{http://www.aclweb.org/anthology/N15-1098}.

\bibitem[{Li et~al.(2010)Li, Lopez-Fernandez, and Veale}]{li2010ucd}
Guofu Li, Alejandra Lopez-Fernandez, and Tony Veale. 2010.
\newblock Ucd-goggle: A hybrid system for noun compound paraphrasing.
\newblock In {\em Proceedings of the 5th International Workshop on Semantic
  Evaluation\/}. Association for Computational Linguistics, pages 230--233.

\bibitem[{Mallinson et~al.(2017)Mallinson, Sennrich, and
  Lapata}]{mallinson-sennrich-lapata:2017:EACLlong}
Jonathan Mallinson, Rico Sennrich, and Mirella Lapata. 2017.
\newblock Paraphrasing revisited with neural machine translation.
\newblock In {\em Proceedings of the 15th Conference of the European Chapter of
  the Association for Computational Linguistics: Volume 1, Long Papers\/}.
  Association for Computational Linguistics, Valencia, Spain, pages 881--893.

\bibitem[{Mitchell and Lapata(2010)}]{mitchell2010composition}
Jeff Mitchell and Mirella Lapata. 2010.
\newblock Composition in distributional models of semantics.
\newblock {\em Cognitive science\/} 34(8):1388--1429.

\bibitem[{Nakov(2013)}]{nakov2013interpretation}
Preslav Nakov. 2013.
\newblock On the interpretation of noun compounds: Syntax, semantics, and
  entailment.
\newblock {\em Natural Language Engineering\/} 19(03):291--330.

\bibitem[{Nakov and Hearst(2006)}]{nakov2006using}
Preslav Nakov and Marti Hearst. 2006.
\newblock Using verbs to characterize noun-noun relations.
\newblock In {\em International Conference on Artificial Intelligence:
  Methodology, Systems, and Applications\/}. Springer, pages 233--244.

\bibitem[{Nastase and Szpakowicz(2003)}]{nastase2003exploring}
Vivi Nastase and Stan Szpakowicz. 2003.
\newblock Exploring noun-modifier semantic relations.
\newblock In {\em Fifth international workshop on computational semantics
  (IWCS-5)\/}. pages 285--301.

\bibitem[{Nesterov(1983)}]{nesterov1983method}
Yurii Nesterov. 1983.
\newblock A method of solving a convex programming problem with convergence
  rate o (1/k2).
\newblock In {\em Soviet Mathematics Doklady\/}. volume~27, pages 372--376.

\bibitem[{Neubig et~al.(2017)Neubig, Dyer, Goldberg, Matthews, Ammar,
  Anastasopoulos, Ballesteros, Chiang, Clothiaux, Cohn
  et~al.}]{neubig2017dynet}
Graham Neubig, Chris Dyer, Yoav Goldberg, Austin Matthews, Waleed Ammar,
  Antonios Anastasopoulos, Miguel Ballesteros, David Chiang, Daniel Clothiaux,
  Trevor Cohn, et~al. 2017.
\newblock Dynet: The dynamic neural network toolkit.
\newblock {\em arXiv preprint arXiv:1701.03980\/} .

\bibitem[{Nulty and Costello(2010)}]{nulty2010ucd}
Paul Nulty and Fintan Costello. 2010.
\newblock Ucd-pn: Selecting general paraphrases using conditional probability.
\newblock In {\em Proceedings of the 5th International Workshop on Semantic
  Evaluation\/}. Association for Computational Linguistics, pages 234--237.

\bibitem[{\'{O}~S\'{e}aghdha and
  Copestake(2009)}]{oseaghdha-copestake:2009:EACL}
Diarmuid \'{O}~S\'{e}aghdha and Ann Copestake. 2009.
\newblock \href{http://www.aclweb.org/anthology/E09-1071}{Using lexical and
  relational similarity to classify semantic relations}.
\newblock In {\em Proceedings of the 12th Conference of the European Chapter of
  the ACL (EACL 2009)\/}. Association for Computational Linguistics, Athens,
  Greece, pages 621--629.
\newblock
  \href{http://www.aclweb.org/anthology/E09-1071}{http://www.aclweb.org/anthology/E09-1071}.

\bibitem[{Pal and Mausam(2016)}]{pal-:2016:W16-13}
Harinder Pal and Mausam. 2016.
\newblock \href{http://www.aclweb.org/anthology/W16-1307}{Demonyms and compound
  relational nouns in nominal open ie}.
\newblock In {\em Proceedings of the 5th Workshop on Automated Knowledge Base
  Construction\/}. Association for Computational Linguistics, San Diego, CA,
  pages 35--39.
\newblock
  \href{http://www.aclweb.org/anthology/W16-1307}{http://www.aclweb.org/anthology/W16-1307}.

\bibitem[{Pantel et~al.(2007)Pantel, Bhagat, Coppola, Chklovski, and
  Hovy}]{pantel-EtAl:2007:main}
Patrick Pantel, Rahul Bhagat, Bonaventura Coppola, Timothy Chklovski, and
  Eduard Hovy. 2007.
\newblock \href{http://www.aclweb.org/anthology/N/N07/N07-1071}{{ISP}: Learning
  inferential selectional preferences}.
\newblock In {\em Human Language Technologies 2007: The Conference of the North
  American Chapter of the Association for Computational Linguistics;
  Proceedings of the Main Conference\/}. Association for Computational
  Linguistics, Rochester, New York, pages 564--571.
\newblock
  \href{http://www.aclweb.org/anthology/N/N07/N07-1071}{http://www.aclweb.org/anthology/N/N07/N07-1071}.

\bibitem[{Pasca(2015)}]{N15-1037}
Marius Pasca. 2015.
\newblock \href{https://doi.org/10.3115/v1/N15-1037}{Interpreting compound noun
  phrases using web search queries}.
\newblock In {\em Proceedings of the 2015 Conference of the North American
  Chapter of the Association for Computational Linguistics: Human Language
  Technologies\/}. Association for Computational Linguistics, pages 335--344.
\newblock
  \href{https://doi.org/10.3115/v1/N15-1037}{https://doi.org/10.3115/v1/N15-1037}.

\bibitem[{Pavlick and Pasca(2017)}]{pavlick-pasca:2017:Long}
Ellie Pavlick and Marius Pasca. 2017.
\newblock \href{http://aclweb.org/anthology/P17-1192}{Identifying 1950s
  american jazz musicians: Fine-grained isa extraction via modifier
  composition}.
\newblock In {\em Proceedings of the 55th Annual Meeting of the Association for
  Computational Linguistics (Volume 1: Long Papers)\/}. Association for
  Computational Linguistics, Vancouver, Canada, pages 2099--2109.
\newblock
  \href{http://aclweb.org/anthology/P17-1192}{http://aclweb.org/anthology/P17-1192}.

\bibitem[{Pedregosa et~al.(2011)Pedregosa, Varoquaux, Gramfort, Michel,
  Thirion, Grisel, Blondel, Prettenhofer, Weiss, Dubourg, Vanderplas, Passos,
  Cournapeau, Brucher, Perrot, and Duchesnay}]{scikit-learn}
F.~Pedregosa, G.~Varoquaux, A.~Gramfort, V.~Michel, B.~Thirion, O.~Grisel,
  M.~Blondel, P.~Prettenhofer, R.~Weiss, V.~Dubourg, J.~Vanderplas, A.~Passos,
  D.~Cournapeau, M.~Brucher, M.~Perrot, and E.~Duchesnay. 2011.
\newblock Scikit-learn: Machine learning in {P}ython.
\newblock {\em Journal of Machine Learning Research\/} 12:2825--2830.

\bibitem[{Pennington et~al.(2014)Pennington, Socher, and
  Manning}]{pennington-socher-manning:2014:EMNLP2014}
Jeffrey Pennington, Richard Socher, and Christopher Manning. 2014.
\newblock \href{http://www.aclweb.org/anthology/D14-1162}{Glove: Global vectors
  for word representation}.
\newblock In {\em Proceedings of the 2014 Conference on Empirical Methods in
  Natural Language Processing (EMNLP)\/}. Association for Computational
  Linguistics, Doha, Qatar, pages 1532--1543.
\newblock
  \href{http://www.aclweb.org/anthology/D14-1162}{http://www.aclweb.org/anthology/D14-1162}.

\bibitem[{Reddy et~al.(2011)Reddy, McCarthy, and
  Manandhar}]{reddy-mccarthy-manandhar:2011:IJCNLP-2011}
Siva Reddy, Diana McCarthy, and Suresh Manandhar. 2011.
\newblock \href{http://www.aclweb.org/anthology/I11-1024}{An empirical study on
  compositionality in compound nouns}.
\newblock In {\em Proceedings of 5th International Joint Conference on Natural
  Language Processing\/}. Asian Federation of Natural Language Processing,
  Chiang Mai, Thailand, pages 210--218.
\newblock
  \href{http://www.aclweb.org/anthology/I11-1024}{http://www.aclweb.org/anthology/I11-1024}.

\bibitem[{Shwartz and Waterson(2018)}]{nc_naacl2018}
Vered Shwartz and Chris Waterson. 2018.
\newblock Olive oil is made of olives, baby oil is made for babies:
  Interpreting noun compounds using paraphrases in a neural model.
\newblock In {\em The 16th Annual Conference of the North American Chapter of
  the Association for Computational Linguistics: Human Language Technologies
  (NAACL-HLT)\/}. New Orleans, Louisiana.

\bibitem[{Socher et~al.(2012)Socher, Huval, Manning, and Ng}]{D12-1110}
Richard Socher, Brody Huval, D.~Christopher Manning, and Y.~Andrew Ng. 2012.
\newblock \href{http://aclweb.org/anthology/D12-1110}{Semantic compositionality
  through recursive matrix-vector spaces}.
\newblock In {\em Proceedings of the 2012 Joint Conference on Empirical Methods
  in Natural Language Processing and Computational Natural Language
  Learning\/}. Association for Computational Linguistics, pages 1201--1211.
\newblock
  \href{http://aclweb.org/anthology/D12-1110}{http://aclweb.org/anthology/D12-1110}.

\bibitem[{Surtani et~al.(2013)Surtani, Batra, Ghosh, and
  Paul}]{surtani2013iiit}
Nitesh Surtani, Arpita Batra, Urmi Ghosh, and Soma Paul. 2013.
\newblock Iiit-h: A corpus-driven co-occurrence based probabilistic model for
  noun compound paraphrasing.
\newblock In {\em Second Joint Conference on Lexical and Computational
  Semantics (* SEM), Volume 2: Proceedings of the Seventh International
  Workshop on Semantic Evaluation (SemEval 2013)\/}. volume~2, pages 153--157.

\bibitem[{Tratz(2011)}]{tratz2011semantically}
Stephen Tratz. 2011.
\newblock {\em Semantically-enriched parsing for natural language
  understanding\/}.
\newblock University of Southern California.

\bibitem[{Tratz and Hovy(2010)}]{tratz-hovy:2010:ACL}
Stephen Tratz and Eduard Hovy. 2010.
\newblock \href{http://www.aclweb.org/anthology/P10-1070}{A taxonomy, dataset,
  and classifier for automatic noun compound interpretation}.
\newblock In {\em Proceedings of the 48th Annual Meeting of the Association for
  Computational Linguistics\/}. Association for Computational Linguistics,
  Uppsala, Sweden, pages 678--687.
\newblock
  \href{http://www.aclweb.org/anthology/P10-1070}{http://www.aclweb.org/anthology/P10-1070}.

\bibitem[{Van~de Cruys et~al.(2013)Van~de Cruys, Afantenos, and
  Muller}]{vandecruys-afantenos-muller:2013:SemEval-20132}
Tim Van~de Cruys, Stergos Afantenos, and Philippe Muller. 2013.
\newblock \href{http://www.aclweb.org/anthology/S13-2026}{Melodi: A supervised
  distributional approach for free paraphrasing of noun compounds}.
\newblock In {\em Second Joint Conference on Lexical and Computational
  Semantics (*SEM), Volume 2: Proceedings of the Seventh International Workshop
  on Semantic Evaluation (SemEval 2013)\/}. Association for Computational
  Linguistics, Atlanta, Georgia, USA, pages 144--147.
\newblock
  \href{http://www.aclweb.org/anthology/S13-2026}{http://www.aclweb.org/anthology/S13-2026}.

\bibitem[{Van Der~Maaten(2014)}]{van2014accelerating}
Laurens Van Der~Maaten. 2014.
\newblock Accelerating t-sne using tree-based algorithms.
\newblock {\em Journal of machine learning research\/} 15(1):3221--3245.

\bibitem[{Versley(2013)}]{versley2013sfs}
Yannick Versley. 2013.
\newblock Sfs-tue: Compound paraphrasing with a language model and
  discriminative reranking.
\newblock In {\em Second Joint Conference on Lexical and Computational
  Semantics (* SEM), Volume 2: Proceedings of the Seventh International
  Workshop on Semantic Evaluation (SemEval 2013)\/}. volume~2, pages 148--152.

\bibitem[{Wubben(2010)}]{wubben2010uvt}
Sander Wubben. 2010.
\newblock Uvt: Memory-based pairwise ranking of paraphrasing verbs.
\newblock In {\em Proceedings of the 5th International Workshop on Semantic
  Evaluation\/}. Association for Computational Linguistics, pages 260--263.

\bibitem[{Xavier and Lima(2014)}]{XAVIER14.125}
Clarissa Xavier and Vera Lima. 2014.
\newblock Boosting open information extraction with noun-based relations.
\newblock In Nicoletta Calzolari~(Conference Chair), Khalid Choukri, Thierry
  Declerck, Hrafn Loftsson, Bente Maegaard, Joseph Mariani, Asuncion Moreno,
  Jan Odijk, and Stelios Piperidis, editors, {\em Proceedings of the Ninth
  International Conference on Language Resources and Evaluation (LREC'14)\/}.
  European Language Resources Association (ELRA), Reykjavik, Iceland.

\bibitem[{Zanzotto et~al.(2010)Zanzotto, Korkontzelos, Fallucchi, and
  Manandhar}]{zanzotto2010estimating}
Fabio~Massimo Zanzotto, Ioannis Korkontzelos, Francesca Fallucchi, and Suresh
  Manandhar. 2010.
\newblock Estimating linear models for compositional distributional semantics.
\newblock In {\em Proceedings of the 23rd International Conference on
  Computational Linguistics\/}. Association for Computational Linguistics,
  pages 1263--1271.

\end{thebibliography}
